\documentclass[10pt,twocolumn]{article} 
\usepackage{simpleConference}
\usepackage{times}
\usepackage{graphicx}
\usepackage{amssymb}
\usepackage{url,hyperref}
\usepackage{graphicx}
\usepackage{caption}
\usepackage{makecell}
\usepackage{booktabs}
\usepackage{mathtools}
\usepackage{float}
\usepackage{array}
\usepackage{multirow}
\usepackage{tabularx}
\usepackage{colortbl}
\usepackage{hyperref}
\usepackage{hhline}
\usepackage[table]{xcolor}
\usepackage{array,hhline}
\RequirePackage[table]{xcolor}
\usepackage{array,hhline}
\usepackage{marvosym}
\usepackage{pgf}
\usepackage{hyperref}
\usepackage{indentfirst}
\usepackage{threeparttable}
\usepackage{amsmath}

\hypersetup{
    colorlinks=true,
    linkcolor=blue,
    filecolor=magenta,      
    urlcolor=cyan,
}
\usepackage{authblk}

\begin{document}

\title{Multi-Level Contextual Network for Biomedical Image Segmentation}

\author{ Amirhossein Dadashzadeh }
\author{ Alireza Tavakoli Targhi}

\affil{ Department of Computer Science, Shahid Beheshti University, Tehran, Iran}
\affil{\small\texttt{a.dadashzade@mail.sbu.ac.ir, a\_tavakoli@sbu.ac.ir}}

\maketitle
\thispagestyle{empty}

\begin{abstract}
Accurate and reliable image segmentation is an essential part of biomedical image analysis. In this paper, we consider the problem of biomedical image segmentation
using deep convolutional neural networks. We propose a new end-to-end network architecture that effectively integrates local and global contextual patterns of histologic primitives to obtain a more reliable segmentation result. Specifically, we introduce a deep fully convolution residual network with a new skip connection strategy to control the contextual information passed forward.
 Moreover, our trained model is also computationally inexpensive due to its small number of network parameters.
We evaluate our method on two public datasets for epithelium segmentation and tubule segmentation tasks.
Our experimental results show that the proposed method provides a fast and effective way of producing a pixel-wise dense prediction of biomedical images. 
\end{abstract}

\section{Introduction} \label{intro}
\begin{figure}[t]

\centering
\includegraphics[scale=0.471]{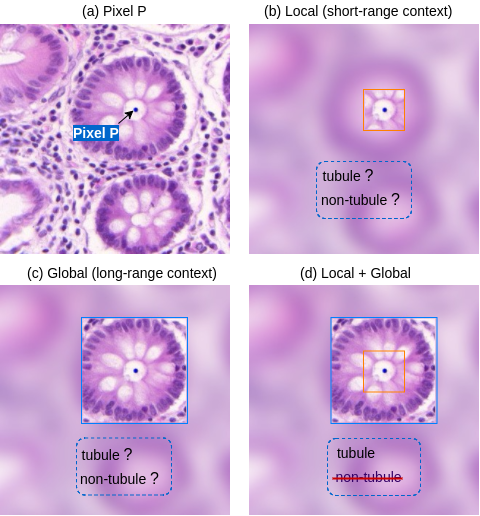}
\caption{Illustration of different contextual information. In this case, the short-range context focuses more on the local information of pixel P and could not capture useful features. While the long-range context is responsible for global region consistency. To aggregate rich contexts, this method combines contextual information from different scales. 
}
\label{f1}
\end{figure}
Analysis and interpretation of stained tissue on microscope slide images is one of the main tools for diagnosis and grading of cancer, which is done manually by a pathologist.
With the progress of machine learning and computer vision techniques, computer-assisted diagnosis (CAD) has been widely used to build tools for the accurate and automatic analysis of these complex and informative image data.

Recently, deep convolutional neural networks (CNNs), have been successfully used to a wide variety of visual recognition tasks such as image classification \cite{r1,r2}, and image segmentation \cite{r3,r8}. Unlike the traditional approach of hand-crafted feature extraction methods \cite{r21,r22,r23}, CNNs learn useful features directly by the composition of multiple linear and non-linear transformations of the training data. Hence, CNN-based methods have gained great popularity for medical image analysis and in particular on biomedical image segmentation tasks \cite{r14,r24,r25}.
In this way, one of the earliest paper in biomedical image segmentation based on CNN was published by Ciresan et al. \cite{r4}. They trained a CNN using the sliding-window technique to predict the class label of each pixel. The main drawback of this model is that the network could lead to storage overhead and is quite slow if we process a high-resolution image. Some studies also used patch-based classification approaches \cite{r5,r6}. For instance, Janowczyk et al. \cite{r6} utilized AlexNet architecture \cite{r2} for patch-based segmentation for histologic primitives (e.g., nuclei, mitosis, tubules, epithelium, etc.). 
However, most recent papers now are based on the Fully Convolution Network (FCN) architecture \cite{r3}. The FCN-like networks remove fully-connected layers to create an efficient end-to-end learning algorithm without extracting the region proposals. These kinds of networks significantly reduce the redundant computations in the original CNN, which can be an advantage in contrast to previous approaches.

U-net \cite{r7}, an FCN based network, is the most well-known architecture in biomedical image segmentation.
The architecture of the U-net consists of the contracting encoder part to capture context and a successive expanding decoder part to enable precise localization.
Segnet \cite{r8} is another popular network proposed for natural image semantic segmentation. Segnet has a similar architecture to U-net but with some differences.
This model reuses the pooling indices from the encoder and learns extra convolutional layers, while U-net adds skip connections from the encoder features to the corresponding decoder activations.
These encoder-decoder based architectures directly concatenate feature maps from earlier layers to recover image detail. However, this strategy is not able to model long-range context, which can be crucial for tasks such as segmentation of tubules in histology images (see Fig. \ref{f1}). Moreover, these models perform complicate decoder module which usually needs substantial computing resources. To address this problem, in this paper we propose a deep multi-level contextual network, a CNN-based architecture that effectively integrates multi-level long-range (global) and short-range (local) contextual information to achieve a more reliable segmentation result without causing too much computation cost. 

In summary, there are three main contributions in our paper which are
as follows:

\begin{description}
\item[$\bullet$] We take advantage of residual learning technique, proposed by \cite{r9}, to build a deeper FCN-like network with fewer parameters and powerful representational ability.
\item[$\bullet$]
We propose a new skip connection strategy using pyramid dilated convolution (PDC) to encode multi-scale and multi-level contextual information passed forward.

\item[$\bullet$]We perform extensive experiments on two digital pathology tasks including epithelium and tubule segmentation \cite{r6} to demonstrate the effectiveness of the proposed model.

\end{description}

The rest of this paper is organized in the following manner. In Section \ref{s3} we explain the methodology of our model. The experimental results are discussed in Section \ref{s4} and finally the conclusion is given in Section \ref{s5}.

\section{Method}\label{s3}
In this section, we start by introducing our fully convolutional residual neural
network, and the pyramid dilated convolution (PDC) module. Then, we describe the overall framework of our deep multi-level contextual network for accurate biomedical image segmentation. 
\subsection{Fully Convolutional Residual Network}
\label{s31}
\begin{figure*}[h]
\centering
\includegraphics[scale=0.49]{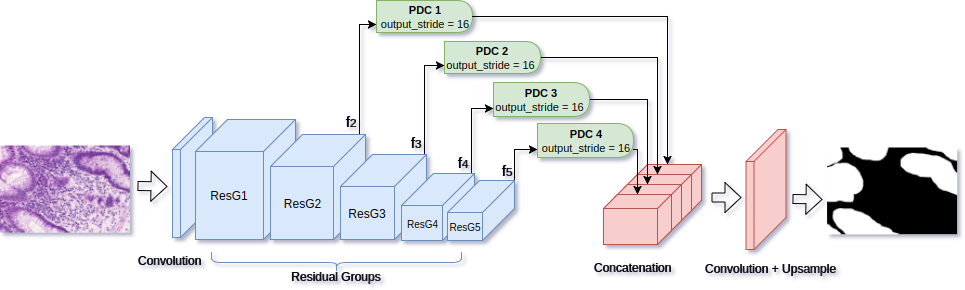}
\caption{Overview of our network framework. A fully convolutional residual network with one $7\times7$ convolution layer and five residual groups is used to compute low and high-level features from the input image. The extracted feature maps ($f_{2},f_{3}, f_{4}, f_{5}$) are then fed to the PDC module to generate multi-level and multi-scale contextual information. Finally, after a concatenation unit, we applied $1\times1$ convolution with a standard 16x bilinear upsampling to build an end-to-end network for pixel-wise dense prediction.
}

\label{f4}
\end{figure*}

Recently, lots of works demonstrated that network depth is of crucial importance in neural network architectures \cite{r11}. However, it is difficult to train deeper networks due to problems such as the risk of overfitting, and difficulty in optimization. Also, the gradient would vanish more easily in such networks. To overcome these problems, He et al. \cite{r9} proposed residual learning technique to ease the training of networks and enables them to be substantially deeper. This technique allows gradients to flow across multiple layers during the training process by using an identity mappings as the skip connections. In this work, we take advantage of the residual learning technique to make a deep encoder network with 46 convolution layers. For this purpose, as shown in Fig. \ref{f4},  we consider 5 residual groups that each group contains 3 residual units which are stacked together. The structure of these units is shown in Fig. \ref{f2}. Each unit can be defined as a general form:

\begin{figure}[t]
\centering
\includegraphics[scale=0.51]{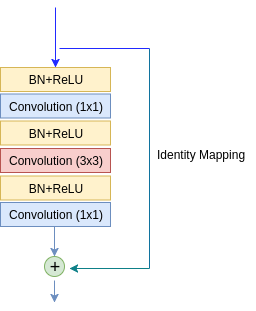}
\caption{The structure of residual unit with bottleneck used in the proposed method.} 
\label{f2}
\end{figure}

\begin{figure*}[h]
\centering
\includegraphics[scale=0.5]{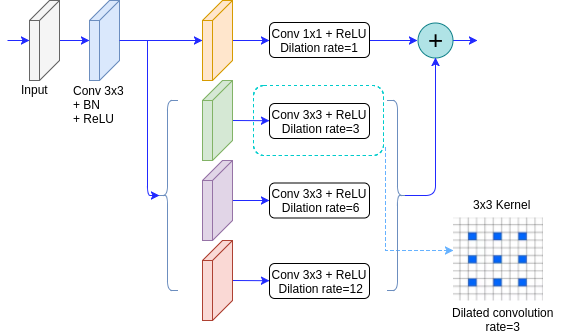}
\caption{Pyramid Dilated Convolution (PDC) module structure.}  \label{f3}
\end{figure*}

\begin{equation} 
  y_{k}=F(x_{k};W)+x_{k}
\end{equation}

Here $x_{k}$ and $y_{k}$ represent the input and output of the $k-th$ unit, and $F(.)$ is the residual function.
As it is clear in Fig. \ref{f2}, we use a bottleneck architecture for each the residual unit, using two $1\times1$  convolution layers. This architecture is responsible for reducing the dimensionality and then restoring it \cite{r9}. 
Also, the downsampling operation is performed by the first $1\times1$ convolution with a stride of 2.

\subsection{Pyramid Dilated Convolution}
\label{s32}
Contextual information has been shown to be extremely important for semantic segmentaton tasks \cite{r12,r13}.
In a CNN, the size of the receptive field can roughly determine how much we need to capture context. Although the wider receptive filed allows us to gather more context, Zhou et al. \cite{r15} showed that the actual size of the receptive fields in a CNN is much smaller than the theoretical size, especially on high-level layers.
To solve this problem, Fisher et al. \cite{r16} propose dilated convolution which can exponentially enlarge receptive field to capture long-range context without losing spatial resolution and increasing the number of parameters.

In Fig. \ref{f1} we show the importance of long-range context in biomedical image segmentation. As it is evidence, the local features only collect information around pixel P (Fig. \ref{f1}(b)). However, capturing more context with a long-range receptive field will bring stronger features which can help to eliminate ambiguity and improve the classification performance.

In this work, we combine different scales of contextual information using a module called pyramid dilated convolution (PDC). This kind of module has been employed successfully in state-of-the-art DeepLabv3 \cite{r10} for semantic segmentation. The PDC module used in this paper consists of four parallel dilated convolutions with different dilated rates. Also, for the first layer of this module, we apply a $3\times3$ convolution layer with stride $s$ to control the resolution of the input feature maps. The refined feature maps of different dilated convolutions are finally concatenated together. Furthermore, The output size of the final feature maps from PDC is 1/16 of the input image. The final module structure is shown in Fig. \ref{f3}.

\subsection{Overall Framework} \label{s33}
With fully convolutional residual network and proposed pyramid dilated convolution (PDC) module, we introduce the overall framework of our network as Fig \ref{f4}. 
In this way, we follow the prior works based on an encoder-decoder architecture \cite{r3,r7,r8} which applied skip connections in the decoder layers to recover the finer information lost in the downsampling layers.

Different from previous works that directly link an encoder layer to a decoder layer, we utilize a PDC module to control the information being passed via the skip connections.
This strategy helps to aggregate multi-scale contextual features from multiple levels. As illustrated in Fig. \ref{f4} in the encoder network, for an input image $I$ the 5 feature maps ($f_{1}, f_{2}, f_{3}, f_{4}, f_{5}$) produce by the 5 residual groups. These feature maps except $f_{1}$ feed to the PDC module to generate multi-scale context maps.
Then a fusion unit is used to concatenate these context maps extracted from hierarchical layers.

Since runtime is of concern, we consider a simple decoder architecture with only one $1\times1$ convolution and one upsampling layer to produce the final segmentation result using sigmoid based classification.

\section{Experimental}\label{s4}
In this section, we first briefly introduce the datasets (Sec. \ref{s41}). Then we explain the details for training the network (Sec. \ref{s42}). Finally, we present experimental results for both tasks of tubule and epithelium segmentation (Sec. \ref{s44}).

\subsection{Dataset} \label{s41}

To show the effectiveness of the proposed method, we carry out comprehensive experiments on two publicly available datasets \cite{r6}.

The first dataset is the colorectal cancer histopathology image dataset which consists of 85 images of size $775\times522$  scanned at 40x and contains a total of 795 delineated tubules.

The second dataset is the estrogen receptor positive (ER+) breast cancer (BCa) histopathology image dataset which consists of 42 images of size $1000\times1000$ scanned at 20x and contains a total of 1735 epithelium regions. To make better use of this dataset, we split each image into four non-overlapping image tiles of $500\times500$.

We use the above datasets for the tasks of tubule segmentation and epithelium segmentation respectively. The tubules and epithelium regions are manually annotated across all images by an expert pathologist and utilized to generate a binary segmentation mask. 

We perform all our experiments using 5-fold cross-validation. In each fold, we consider 80\% images for training and 20\% images for testing. Moreover, to improve generalization and prevent overfitting, we apply the strategy of data augmentation to generate more training data. The augmentation transformations include rescaling and flipping horizontally and vertically.
\subsection{Training Network}\label{s42}
The Keras open-source deep learning library \cite{r17} with the backend of tensorflow \cite{r18} is used to implement the proposed method.

In the training phase the cross-entropy loss is applied as the cost functions: 
  \begin{equation} 
  L(t,p)= - \sum t(x)log(p(x))
\end{equation}
where $t$ and $p$ corresponding to prediction and target.
We performed optimization via Adam optimization algorithm \cite{r19}. The learning rate is initialized as 0.001, and $\beta_{1}=0.9$ and $\beta_{2}=0.999$.
We also applied dropout \cite{r2} (with $p=0.5$) before the sigmoid classification layer.
The proposed framework has been trained using an Nvidia GTX 1080TI GPU with 3.00GHz 4-core CPU and 16GB RAM. The training phase takes around 2 hours for each fold. The maximum number of epochs were fixed at 200, and the model with the best validation loss was selected and evaluated. Our trained model takes roughly 35ms for a $480\times320$ input RGB image.

\begin{figure*}[h]
\centering
\includegraphics[scale=0.453]{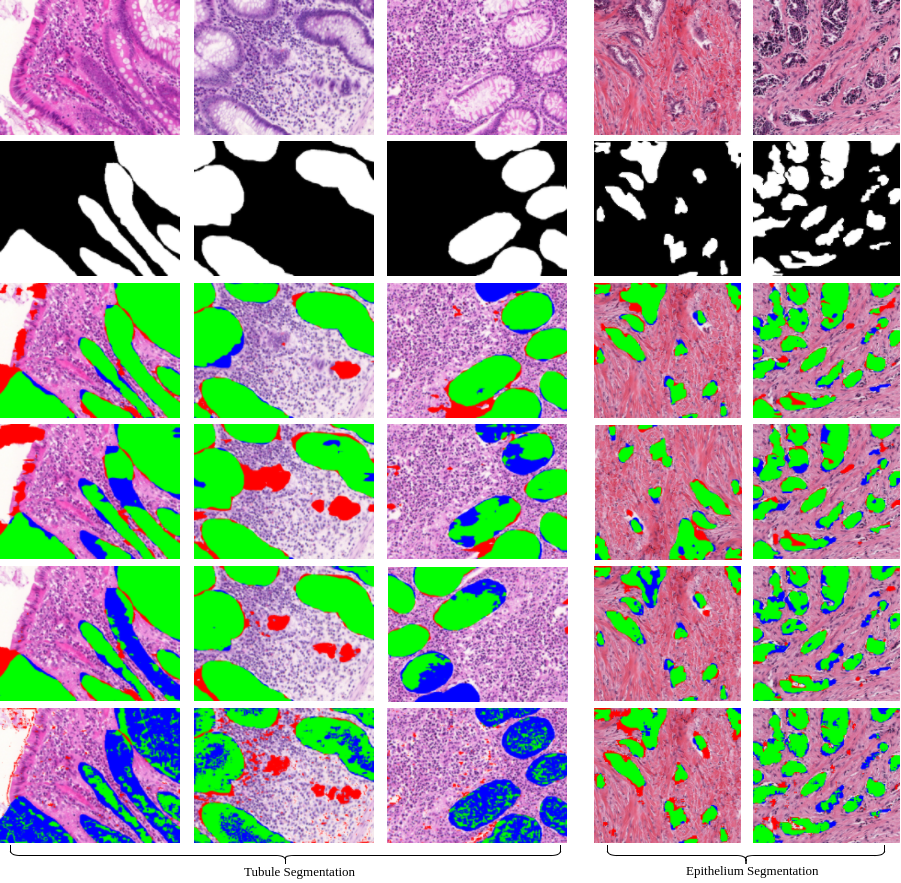}
\caption{Visual comparison on sub-images taken from the test images, (from top to bottom): input image, ground truth, proposed model, baseline model, Segnet \cite{r8} and U-net \cite{r7}. Green pixels show true positive, red false positive and blue false negative. Note that we perfomed data agumentation for all of these models.
} \label{f5}
\end{figure*}

\subsection{Results and Comparisons}\label{s44}
To carry out a quantitative comparison of accuracy of the proposed method, we measure F1-score which is the harmonic mean of the Precision and Recall \cite{r20} with a value in the range of $[0,1]$, defined bellow:
\\
\begin{equation} 
 \text{Precision}=\frac{\text{tp}}{\text{tp}+\text{fp}}
\end{equation}
\begin{equation} 
 \textnormal{Recall}=\frac{\text{tp}}{\text{tp}+\text{fn}}
\end{equation}
\begin{equation} 
  \text{F1\text{-}score}=2\times\frac{\text{Precision}\times \textnormal{Recall}}{\text{Precision}+\textnormal{Recall}}
\end{equation}
\\
where tp is true positive, fp denotes false positive and fn denotes false negative. 

\begin{table}[h]
\footnotesize
\centering
\captionsetup{justification=centering}
\begin{flushleft}
\caption{Quantitative comparison of epithelium segmentation performance on ER+BCa dataset. \lq DA\rq \space
refers to data augmentation we performed. } \label{t4epi}
\end{flushleft}
    \begin{tabular}{{m{0.16\textwidth}m{0.09\textwidth}m{0.09\textwidth}m{0.09\textwidth}m{0.4\textwidth}}} 
        \toprule
        Method & F1-score &  Std.Dev   \\
        \midrule
        Original Paper \cite{r6} & $0.84$ & $-$ \\
        U-net \cite{r7} & $0.8912$ & $\pm  0.01$\\
        Segnet \cite{r8} & $0.8846$ & $\pm0.02$\\

        Baseline & $0.8831$&$\pm0.02$ \\
        Baseline+DA & $0.8981$&$\pm0.02$ \\
        \textbf{Baseline+DA+PDC} & \textbf{0.9066}&$\pm0.01$ \\
  
        \bottomrule
    \end{tabular}
    \label{tab:PPer}
\end{table}
\begin{table}[h]
\footnotesize
\centering
\captionsetup{justification=centering}
\begin{flushleft}
\caption{Quantitative comparison of tubule segmentation performance on colorectal cancer  dataset. \lq DA\rq \space
refers to data augmentation we performed.} \label{t4tub}
\end{flushleft}
    \begin{tabular}{{m{0.16\textwidth}m{0.09\textwidth}m{0.09\textwidth}m{0.09\textwidth}m{0.4\textwidth}}} 
        \toprule
        Method & F1-score &  Std.Dev   \\
        \midrule
        Original Paper \cite{r6} & $0.83$ & $-$ \\
        U-net \cite{r7} & $0.7862$ & $\pm  0.05$\\
        Segnet \cite{r8} & $0.8882$ & $\pm0.02$\\

        Baseline & $0.8242$&$\pm0.03$ \\
        Baseline+DA & $0.8646$&$\pm0.03$ \\
        \textbf{Baseline+DA+PDC} & \textbf{0.8950}&$\pm0.02$ \\
  
        \bottomrule
    \end{tabular}
    \label{tab:PPer}
\end{table}
The evaluation results are obtained for the both tasks of tubule segmentation and epithelium segmentation with the same settings.
Our baseline network is adapted from a fully convolutional residual network with 46 convolution layers which described in Section \ref{s31}.
To illustrate the performance of our multi-level contextual network, we compared the proposed method with our baseline network.
To evaluate the baseline, we applied a $1\times1$ convolution with 16x upsampling layer after the last residual group. We also compared the performance of our method with   three state-of-the-art methods including 1) a patch-based segmentation method proposed in the original paper \cite{r6} and, 2) the original U-net which won
several grand challenges recently and, 3) the original SegNet architecture which is a popular network for natural image semantic segmentation.
The quantitative segmentation results are reported in Table \ref{t4epi} and Table \ref{t4tub}.
All our results except the first row are trained with data augmentation technique.

It is clear that the proposed method achieved the best performance on both of the tasks and outperformed all the other methods, which demonstrates
the significance of the multi-level contextual features.
Note that we did not do any post-processing of the resulting segmentation.

\begin{figure}[h]
\centering
\includegraphics[scale=0.48]{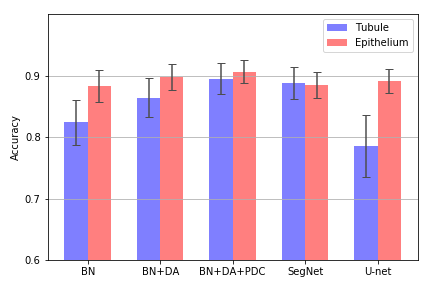}
\caption{Segmentation results (Mean $\pm$ Std.Dev) of different methods on both datasets. \lq BN\rq \space
 denotes our baseline network and \lq DA\rq \space
refers to data augmentation we performed.} 
\label{f8}
\end{figure}

A visualization of the results of the proposed method and comparison with the other methods is shown in Fig. \ref{f5}, As shown in the samples, our segmentation results have fewer false negatives with higher accuracy. In addition, our method is not only more accurate but also is faster due to its fewer parameters as Table \ref{t4parm}. Fig. \ref{f8}  shows the overall segmentation results with better visualization on both datasets.

\begin{table}[h]
\footnotesize
\centering
\captionsetup{justification=centering}
\begin{flushleft}
\caption{Comparison of model size and number of parameters for various deep models. } \label{t4parm}
\end{flushleft}
    \begin{tabular}{{m{0.08\textwidth}m{0.14\textwidth}m{0.13\textwidth}}} 
        \toprule
        Model & Parameters (Million)  & Model size (MB)  \\
        \midrule
        U-net \cite{r7} & \multicolumn{1}{c}{7.76} & \multicolumn{1}{c}{93.3} \\
        Segnet \cite{r8} & \multicolumn{1}{c}{29.45} & \multicolumn{1}{c}{353.7} \\

        Proposed & \multicolumn{1}{c}{\textbf{3.4}} & \multicolumn{1}{c}{\textbf{41.8}} \\
  
        \bottomrule
    \end{tabular}
    \label{tab:PPer}
\end{table}

\section{Conclusion}\label{s5}
In this paper, we address the problem of biomedical image segmentation using a novel end-to-end deep learning framework. Since the contextual information is crucial to obtain good segmentation results, we have presented a deep multi-level contextual network that effectively integrates multi-level contextual features to accurately segment histologic primitives (e.g., nuclei and tubules) from histology images without causing too much computation cost. In this way, we used a deep fully convolution residual network as the encoder network. Then we applied a new skip connection strategy using pyramid dilated convolution modules to modulate multi-scale contextual information passed forward. 
We have evaluated our method on two public available datasets for epithelium segmentation and tubule segmentation tasks.
Our experimental results show that our proposed model is superior to several state-of-the-art methods.

\bibliographystyle{abbrv}

\end{document}